\documentclass[3p,times]{elsarticle}

\usepackage[separator = {\newline}]{credits}
\usepackage[utf8]{inputenc}
\usepackage{graphicx}%
\usepackage{multirow}%
\usepackage{amsmath,amssymb,amsfonts}%
\usepackage{amsthm}%
\usepackage{mathrsfs}%
\usepackage[title]{appendix}%
\usepackage{xcolor}%
\usepackage{textcomp}%
\usepackage{manyfoot}%
\usepackage{comment}
\usepackage{booktabs}%
\usepackage{gensymb}
\usepackage{algorithmicx}%
\usepackage{algpseudocode}%
\usepackage{listings}%
\usepackage{dirtytalk}
\usepackage{hyperref}
\usepackage{pifont}
\usepackage{caption}
\usepackage{tabularx}
\usepackage[ruled,linesnumbered]{algorithm2e}

\usepackage{adjustbox}
\usepackage{float}
\usepackage{tikz}
\usepackage{pgfplots}
\pgfplotsset{compat=1.15}
\usepackage{mathrsfs}
\usetikzlibrary{arrows}
\usepackage{tikz-cd}
\usepackage{tkz-graph}
\usetikzlibrary{decorations.pathmorphing,decorations.markings,arrows,calc,shapes.geometric,arrows.meta,positioning}
\tikzset{math3d/.style=
	{x= {(-0.353cm,-0.353cm)}, z={(0cm,1cm)},y={(1cm,0cm)}}}
\tikzset{JLL3d/.style=
	{x= {(0.4cm,-0.2cm)}, z={(0cm,1cm)},y={(-1cm,0cm)}}}
\usepackage{perpage} 
\MakePerPage{footnote}
\usepackage{framed}
\usepackage{makecell}

\usepackage{longtable}
\definecolor{Chocolat}{rgb}{0.36, 0.2, 0.09}
\definecolor{BleuTresFonce}{rgb}{0.215, 0.215, 0.36}
\definecolor{BleuMinuit}{RGB}{0, 51, 102}

\hypersetup{citecolor=BleuMinuit, linkcolor=Chocolat, filecolor=black, urlcolor=BleuMinuit}
\usepackage{xargs} 
\definecolor{armygreen}{rgb}{0.29, 0.33, 0.13}    

\usepackage[colorinlistoftodos,prependcaption,textsize=small]{todonotes}

\newcommandx{\change}[2][1=]{\todo[linecolor=green,backgroundcolor=green!25,bordercolor=green,#1]{#2}}

\newcommandx{\info}[2][1=]{\todo[linecolor=yellow,backgroundcolor=yellow!25,bordercolor=yellow,#1]{#2}}
\newcommandx{\question}[2][1=]{\todo[linecolor=blue,backgroundcolor=blue!25,bordercolor=blue,#1]{#2}}
\newcommandx{\idea}[2][1=]{\todo[linecolor=orange,backgroundcolor=orange!25,bordercolor=orange,#1]{#2}}
\newcommandx{\link}[2][1=]{\todo[linecolor=black,backgroundcolor=white!25,bordercolor=black,#1]{#2}}
\newcommand{\ts}{\textsuperscript}

\newcommandx{\commentST}[2][1=]{\todo[linecolor=cyan,backgroundcolor=blue!25,bordercolor=cyan,#1]{#2}}

\newcommand{\FedAvg}{\texttt{FedAvg}}
\newcommand{\FedProx}{\texttt{FedProx}}

\def\G_#1{\mathfrak{#1}} 
\def\t_#1{\widetilde{#1}}

\theoremstyle{thmstyleone}%
\theoremstyle{thmstyletwo}%

\theoremstyle{thmstylethree}%

\begin{document}
\let\today\relax
\makeatletter
\def\ps@pprintTitle{%
    \let\@oddhead\@empty
    \let\@evenhead\@empty
    \def\@oddfoot{\footnotesize\itshape
         {} \hfill\today}%
    \let\@evenfoot\@oddfoot
    }
\makeatother

\let\WriteBookmarks\relax
\def\floatpagepagefraction{1}
\def\textpagefraction{.001}

\title{Wind turbine condition monitoring based on intra- and inter-farm federated learning}

\author[biel]{Albin Grataloup\corref{cor1}}

\ead{albin.grataloup@bfh.ch}



\address[biel]{Bern University of Applied Sciences, School of Engineering and Computer Science, Quellgasse 21, 2501 Biel, Switzerland }
                                              

\author[biel,lugano]{Stefan Jonas}


\ead{stefan.jonas@bfh.ch}


\address[lugano]{Università della Svizzera italiana, Faculty of Informatics, Via la Santa 1, 6962 Lugano-Viganello, Switzerland}

\author[biel,deft]{Angela Meyer}

\cortext[cor1]{Corresponding author}



\ead{angela.meyer@bfh.ch}


\address[deft]{Delft University of Technology, Department of Geoscience and Remote Sensing, Stevinweg 1, 2628 Delft, The Netherlands}


\begin{abstract} 
As wind energy adoption is growing, ensuring the efficient operation and maintenance of wind turbines becomes essential for maximizing energy production and minimizing costs and downtime. Many AI applications in wind energy, such as in condition monitoring and power forecasting, may benefit from using operational data not only from individual wind turbines but from multiple turbines and multiple wind farms. Collaborative distributed AI which preserves data privacy holds a strong potential for these applications. Federated learning has emerged as a privacy-preserving distributed machine learning approach in this context. We explore federated learning in wind turbine condition monitoring, specifically for fault detection using normal behaviour models. We investigate various federated learning strategies, including collaboration across different wind farms and turbine models, as well as collaboration restricted to the same wind farm and turbine model. Our case study results indicate that federated learning across multiple wind turbines consistently outperforms models trained on a single turbine, especially when training data is scarce. Moreover, the amount of historical data necessary to train an effective model can be significantly reduced by employing a collaborative federated learning strategy. Finally, our findings show that extending the collaboration to multiple wind farms may result in inferior performance compared to restricting learning within a farm, specifically when faced with statistical heterogeneity and imbalanced datasets.
\end{abstract}

\maketitle

\section{Introduction}  
The deployment of wind turbines for renewable energy generation is witnessing exponential growth globally \cite{euwindreoprt2023, globalwindreport2023}, driven by the transition towards sustainable energy sources. Ensuring the efficient and reliable operation of wind turbines is critical to maximizing energy production and minimizing downtime and maintenance costs. Condition monitoring and anomaly detection play a pivotal role, offering insights into the health and performance of critical components. Deep learning methods, in particular, have risen as an efficient approach to anomaly detection \cite{DeepLearningforAnomalyDetection:AReview, AnomalyDetectiononWindTurbinesBasedonaDeepLearningAnalysisofVibrationSignals, AnomalyDetectionforWindTurbinesUsingLongShort-TermMemory-BasedVariationalAutoencoderWassersteinGenerationAdversarialNetworkunderSemiSupervisedTraining, AnomalyDetectiononSmallWindTurbineBladesUsingDeepLearningAlgorithms}. However, the demanding data prerequisites of deep learning models present a major challenge as they necessitate either an abundance of labeled data from faulty operation or, in our scenario, a large amount of fault-free data for training a \emph{normal behaviour model} (NBM). A NBM operates by predicting target variables like component temperatures or power output that are crucial for assessing system health or performance. Anomalies are identified when the predicted target variable diverges significantly from the measured value of the target variable, such as detecting an abnormal spike in component temperatures compared to the system's normal operational values. Condition monitoring and anomaly detection are extensively studied fields within the area of wind turbine operations. In recent years, deep learning has emerged as a particularly promising approach for condition-monitoring tasks. Among the methodologies employed, NBMs have gained prominence. These models rely on the comparison between critical features measured in wind turbines and their corresponding predicted values, serving as indicator for assessing wind turbine health \cite{Maron2022, Jonas2023,Anartificialneuralnetwork-basedconditionmonitoringmethodforwindturbineswithapplicationtothemonitoringofthegearbox}.

Training an effective NBM requires a substantial amount of data, which can be time-consuming or even impractical to obtain. For wind turbines, the fastest way to amass sufficient data is by collecting training data from multiple turbines. However, this approach raises significant data privacy concerns as manufacturers and operators are hesitant to share operational data due to strategic business interests \cite{RenewablesSharedataonwindenergy}. A single wind turbine would require a significant amount of time to gather enough data to train a representative and accurate NBM. To address this issue, we propose privacy-preserving collaborative learning methods to leverage training data collected from multiple wind turbines simultaneously. Federated Learning (FL) emerged as a promising paradigm to address these challenges \cite{fl_review_grataloup_jonas_meyer}. FL enables collaborative decentralised model training across multiple wind turbines while preserving their data privacy. By exchanging only FL model parameters and not operational data, the sensitive operation data of each wind turbine remains local and inaccessible to others. This approach allows wind turbines to collaboratively train an effective NBM with less data, without compromising their privacy. 

Federated learning has gained traction across various domains \cite{liReviewApplicationsFederated2020}. It was also adopted in renewable energy sectors \cite{fl_review_grataloup_jonas_meyer}, notably in wind energy applications, for tasks such as wind power forecasting \cite{Windpowerforecastingconsideringdataprivacyprotection, DeepFederatedLearning-BasedPrivacy-PreservingWindPowerForecasting, ACyber-Securegeneralizedsupermodelforwindpowerforecastingbasedondeepfederatedlearningandimageprocessing, APrivacy-preservingWindSpeedPredictionMethodBasedonFederatedDeepLearning, AnefficientfederatedtransferlearningframeworkforcollaborativemonitoringofwindturbinesinIoE-enabledwindfarms}, to obtain significantly more accurate forecasts compared to local models. It also has shown promising results in fault detection applications, exhibiting improved performance over local training methodologies for tasks such as blade icing detection \cite{Windturbinebladeicingdetection:afederatedlearningapproach, AClass-ImbalancedHeterogeneousFederatedLearningModelforDetectingIcingonWindTurbineBlades, HumanKnowledge-basedCompressedFederatedLearningModelforWindTurbineBladeIcingDetection, ABlockchain-EmpoweredCluster-BasedFederatedLearningModelforBladeIcingEstimationonIoT-EnabledWindTurbine}, fault detection \cite{DeepFedWT:Afederateddeeplearningframeworkforfaultdetectionofwindturbines, FederatedMulti-ModelTransferLearningBasedFaultDiagnosiswithPeer-to-PeerNetworkforWindTurbineCluster} and condition monitoring  with an NBM \cite{TowardsFleet-wideSharingofWindTurbineConditionInformationthroughPrivacy-preservingFederatedLearning}. Despite these advancements, the application of FL for training NBMs for wind turbines remains largely unexplored. By investigating collaborative NBM training across multiple wind turbines and wind farms, our study builds upon previous work on the potential of FL for training NBMs \cite{TowardsFleet-wideSharingofWindTurbineConditionInformationthroughPrivacy-preservingFederatedLearning}. We assess the benefits and limits of collaboration particularly by addressing the effect of statistical heterogeneity between different wind turbines and wind farms. We also analyse how FL affects the time required to collect training data for a NBM through collaborative learning of multiple wind turbines.

\begin{figure}[H]
  \centering
  \includegraphics{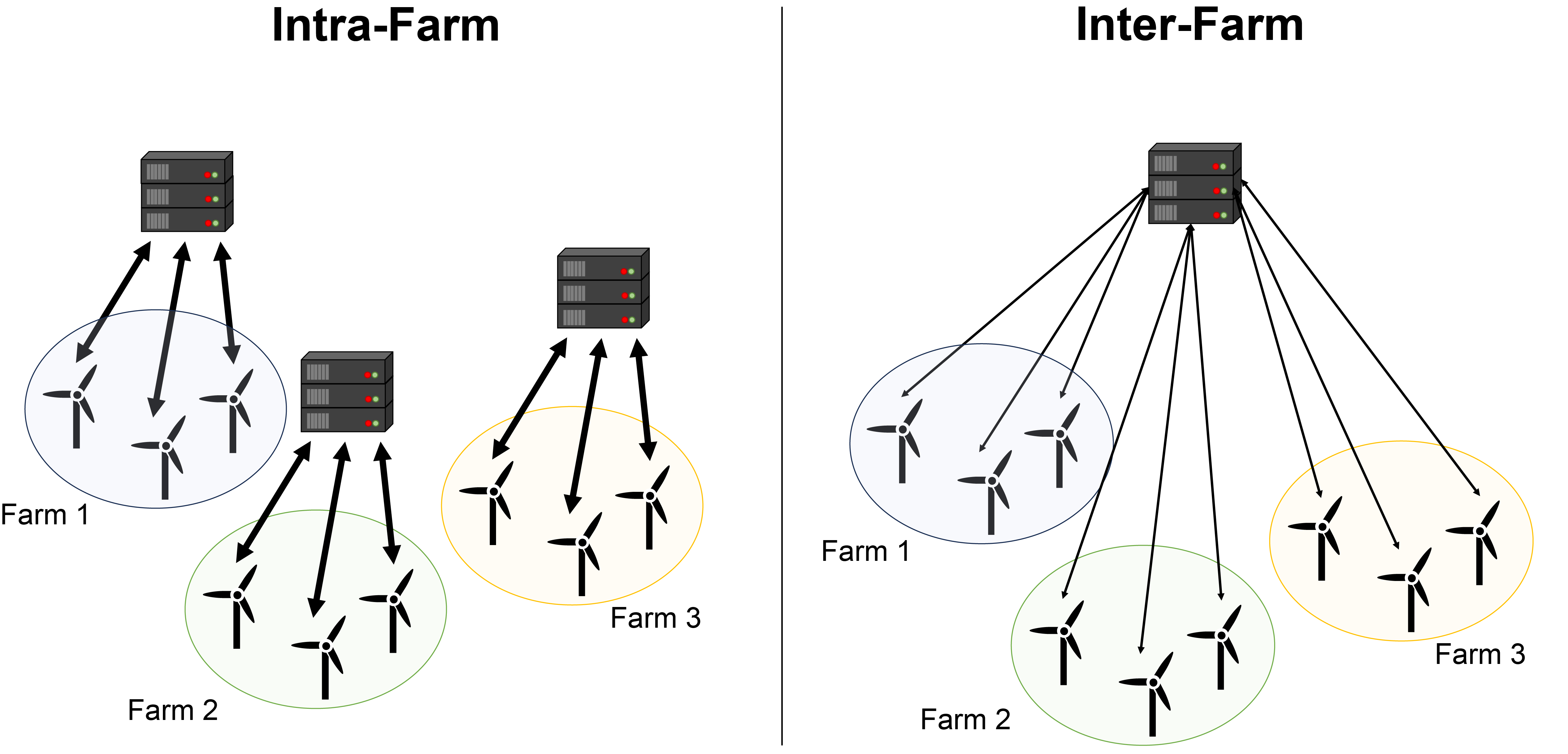}
  \caption{Left: Intra-farm learning, only the turbines from the same wind farms collaborate. Right: Collaborative inter-farm learning on the turbines of all wind farms.}
  \label{fig:inter-farm}
\end{figure}

The objectives of our study are twofold: First, we assess the effectiveness of collaborative federated learning strategies among wind turbines of multiple wind farms, comparing intra- and inter-farm collaborative federated learning (Figure \ref{fig:inter-farm}). Second, we quantify the time savings in collecting training data for a NBM through collaborative FL across multiple wind turbines compared to collecting training data from a single WT (referred to as "local" data). 

\section{Federated learning for condition monitoring of wind turbines}

FL is a collaborative deep-learning framework that involves distributed participants referred to as clients. In our scenario, each wind turbine acts as an individual client and aims to collaboratively train a machine learning model for condition monitoring. In FL, it is crucial that the clients never share their locally stored data in order to preserve data privacy. The iterative FL training process involves that clients train local models, such as a NBM, using only their private local dataset, and then transmit the model parameters to a central server where they are aggregated. This approach ensures the privacy of locally stored client data and can provide a viable solution to overcome data scarcity \cite{fl_review_grataloup_jonas_meyer, TowardsFleet-wideSharingofWindTurbineConditionInformationthroughPrivacy-preservingFederatedLearning}. FL with a central server involves the following iterative steps:

\begin{figure}[H]
  \centering
  \begin{minipage}{.6\linewidth}
    \begin{algorithm}[H]
      \SetAlgoLined
      {R: number of training rounds.}
      \newline Initialization of the server model\;
      
      \For{$t = 1 \cdots R$}{
        - N clients receive a global FL model from the server\;
        
        - All clients independently perform training updates on this model using only their local datasets $\mathscr{D}_i$,\, $i=1 \dots N$;
        
        - The clients send the parameters of their updated local models $\mathscr{M}_i$, $i=1 \dots N$, to the server\;
        
        - The server aggregates the parameters of all models $\mathscr{M}_i$, to obtain the updated global FL model\;
      }
      \caption{Centralised federated learning}
    \end{algorithm}
  \end{minipage}
\end{figure}

The most widely applied FL framework is the Federated Averaging (\texttt{FedAvg}) algorithm \cite{Communication-EfficientLearningofDeepNetworksfromDecentralizedData} in which the aggregation step consists of averaging the received model parameters

\begin{equation}
\omega^{t+1} = \sum\limits_{i=1}^N  \frac{n_i}{n} \omega_i^{t+1}
\label{eq:fedavg}
\end{equation}

 where $\omega^t$ and $\omega_i^t$ denote the global model parameters and the model parameters of client $i$, respectively, in training round $t$. $n_i$ denotes the amount of data available to client $i$ while $n$ is the total amount of available data across all clients involved in the aggregation. 


Although \FedAvg \ has demonstrated empirical success and serves as a cornerstone in many FL algorithms, its effectiveness in real-world applications can be hindered by \emph{statistical heterogeneity}, where the data distributions differ across the clients participating in the learning process. The clients' data may differ in their statistical properties and in size, for example, because of differences in feature distributions or in label distributions. In wind turbines, individual turbines may display differing mechanical characteristics and possibly even differing turbine models and supervisory control and data acquisition (SCADA) systems may be involved. Statistical heterogeneity poses a challenge for FL model training and convergence because the aggregated model must learn to generalize across the diverse datasets. The variations among clients result in differences in the statistical distributions of their local datasets, leading to non-identically distributed (non-iid) data distributions.


Fleets of industrial assets, such as wind turbines, can display significant statistical heterogeneity across clients. 
In such settings, global FL models tend to exhibit suboptimal performance \cite{hsuMeasuringEffectsNonIdentical2019, NonIIDDataSilos} 
compared to locally trained models. The latter may even achieve higher accuracy than their globally trained counterparts \cite{TowardsFleet-wideSharingofWindTurbineConditionInformationthroughPrivacy-preservingFederatedLearning}.
As a result, some clients may lack incentives to participate in training the global FL model (\cite{TowardsFleet-wideSharingofWindTurbineConditionInformationthroughPrivacy-preservingFederatedLearning, AdaptivePFL, SalvagingLocalAdaptation}). To address this challenge, Personalised FL (PFL) has been proposed to customize global FL models to individual clients. PFL retains the advantages of collaborative learning while tailoring the resulting global FL models to each client's specific local data. Various PFL approaches exist, including client clustering \cite{HierarchicalClusteringBriggs, Multi-CenterFederatedLearning:ClientsClusteringforBetterPersonalization, Privacy-preservingknowledgesharingforfewshotbuildingenergyprediction:Afederatedlearningapproach, ObjectiveFunctionClustering}, personalised model layers \cite{PersonalizationLayers}, meta-learning \cite{PFLHyperNetworks}, and fine-tuning methods \cite{FedAvgWithFineTuning}. We refer to \cite{TowardsPFL} and \cite{SurveyPFL} for a comprehensive overview of customisation approaches, and to \cite{fl_review_grataloup_jonas_meyer} for PFL applications in renewable energy contexts.

Training a NBM on data from a single WT requires a significant amount of data representative of the WT's normal operational behaviour, which may not always be available. For example, this is typically the case in newly installed wind farms or after component updates and replacements. The resulting lack of data to train a representative and accurate model is known as the \emph{cold start problem} in computer science, e.g., \cite{Serral2011}. We propose to exploit data gathered from multiple wind turbines to reduce the amount of time required for collecting data for training NBMs. We refer to the time savings as the \emph{cold start speed up} because it is the speed up achieved by training a NBM from scratch through collaborative training rather than training on only local data. Due to privacy considerations, the data from individual turbines are kept confidential, so no data sharing with other wind turbines or servers is allowed. We employ FL for collaborative learning across different wind turbines and different wind farms. We assess the impact of having multiple turbines with different specifications in different wind farms on the efficacy of collaborative learning.
Condition monitoring often relies on NBMs which simulate the normal operation behaviour of the monitored WT components under the current environmental and operation conditions. NBMs are trained on WT data from fault-free operation periods, and allow quantifying the deviations between the measured target variables and their expected values as simulated by the NBM. 
We consider identical feature and label spaces of all client wind turbines in this study, so the same SCADA variables are used as features and target variables of the NBMs of all WTs and wind farms considered in this study. Other types of FL, such as vertical FL and federated transfer learning \cite{FederatedTransferLearning} are not considered in this study.

\section{Intra- and inter-farm federated learning of normal behaviour models}

\subsection{Wind farm data}
\label{sec:data-description}
We investigate FL for wind turbine condition monitoring using SCADA data of WTs from three different wind farms (Table \ref{tab:wind_farm}). The wind farms provide different WT models and site conditions, which can give rise to statistical heterogeneity of the WTs' SCADA variables.
The wind farms Penmanshiel and Kelmarsh exhibit similar configurations, sharing identical SCADA variables, whereas the EDP wind farm features a different SCADA system. We chose 10-minute averages of wind speed, ambient temperature, and wind direction as input features for the NBM, with gear-bearing temperature as the target variable to be predicted. The SCADA data were cleaned by removing curtailment periods and outliers. Wind speed and ambient temperature were normalised, while wind direction was cyclically encoded by a sine-cosine transformation.
The SCADA datasets of the EDP, Kelmarsh and Penmanshiel wind farms contain significant fractions of missing values of 1\%, 5\%, and 30\%, respectively. The FL algorithm \FedAvg \ applied in our study weighs the WTs' contribution to the training in accordance with the fraction of training data available from them (eq. \ref{eq:fedavg}), so data imbalance can affect the learning process in intra- and inter-farm FL.

\begin{table}[h!]
\centering
    \begin{tabular}{|l|l|l|l|}
    \hline
        \textbf{Wind farm}& \textbf{Penmanshiel} & \textbf{Kelmarsh} & \textbf{EDP}\\
        \hline
        \textbf{Location} & 55.906\degree N, 2.262\degree W & 52.402\degree N, 0.945\degree W & unknown\\
        \hline
        \textbf{Turbine model} & Senvion MM82 & Senvion MM92 & unknown\\
         \hline
       \textbf{Time period} & 2016–2021 & 2016–2021 & 2016–2017 \\
         \hline
        \textbf{Number of turbines} & 14 & 6 & 4\\
         \hline
      \textbf{Rated power}& 2.05 MW & 2.05 MW & 0.34 MW\\
         \hline
       \textbf{Rotor diameter} & 82 m & 92 m &  unknown\\
         \hline
         \textbf{Cut-out wind speed} & 25 m/s & 24 m/s & 25 m/s\\
         \hline
         \textbf{SCADA data source} & \cite{penmanshiel} & \cite{kelmarsh} & \cite{edp2016} \& \cite{edp2017} \\
         \hline
    \end{tabular}
    \caption{Wind farms that provided SCADA data to our study}
    \label{tab:wind_farm}
\end{table}

\subsection{Model architecture and training}

The NBM trained in this study is an LSTM (Long Short-Term Memory) network consisting of layers of LSTM units followed by fully connected layers. The LSTM layers capture temporal dependencies in the data. We utilized a 24-hour trailing window to make sure the model can capture the operational conditions of the past 24 hours. The NBM predicts the expected gear-bearing temperature at the end of this time window. 
Hyperparameters of the LSTM network were optimised for a single randomly selected turbine from the Penmanshiel wind farm whose data presented greater complexity than the turbines of other wind farms. The resulting LSTM model comprises two LSTM layers of sizes 16 and 64, respectively, and ReLU activation, followed by two fully connected layers of sizes 64 and 32 with ReLU activation. For this case study, we randomly selected four turbines from each of the three wind farms to reduce data imbalance and computational cost for FL model training. SCADA data from the resulting 12 WTs was used to train a NBM with FL. 

To assess the reduction of the training data accumulation time (cold start speed up), we train our NBM using increasing time intervals of training data. We will start by selecting a specific date as the start date. Then, for each start date, we will train multiple NBMs, each incorporating progressively more historical data by using different time ranges of training data. The time ranges commence from the selected start date and incrementally increase by one week, reaching a maximum of 12 weeks. This approach allows us to evaluate the impact of FL on reducing the time required to accumulate adequate data for model training.
To account for seasonal variations and avoid bias towards any particular season, we select four different start dates spread evenly throughout the year: December 2016, March 2017, June 2017, and September 2017. For each of these start dates, we train models using all 12 different time ranges, enabling a comprehensive evaluation of the cold start speed up achieved by FL across different seasons. For each start date, a test set is given by the 4-week time window that follows the 12\ts{th} week of training data. 
Each local dataset is split into 80\% for training and 20\% for validation. We note that in the case of only one or up to two weeks of data, this may result in a validation dataset that is not fully independent from the training dataset due to autocorrelation of environmental condition time series.

For each training set described above (defined by a selected start date and time range), we trained normal behaviour models of gear bearing temperature according to three different learning strategies representing different types of collaboration:
\begin{itemize}
    \item \emph{Local learning}: Each wind turbine independently trains its own model using only its local data without any collaboration with other turbines.
    \item \emph{Intra-farm learning}:  Each wind farm utilizes \FedAvg \ to train its own FL model, with no exchange of data or knowledge between different wind farms. In intra-farm learning, FL models are trained on similar WT clients, as turbines within a given wind farm typically exhibit similar and correlated SCADA data patterns.
    \item \emph{Inter-farm learning}: Turbines of multiple wind farms participate in a single \FedAvg \ learning process. The participating wind turbines involve different WT models, SCADA systems, and geographic locations, which results in significant data heterogeneity among the participating WT clients.
\end{itemize}
 
The intra- and inter-farm learning strategies are illustrated in Figure \ref{fig:inter-farm}. For each FL strategy, we assessed the performance of the trained normal behaviour models with and without fine-tuning. Fine-tuning involves retraining the global model on each wind turbine's local training data after the FL training process. We provide our implementation on GitHub \footnote{Code available at \protect\url{https://github.com/EnergyWeatherAI/FL-Wind-NBM}}.

\section{Results and discussion} 
\label{sec:results}
\subsection{FL outperforms local training if training data are limited} 
\label{sec:outperformance}
We compare the FL strategies by analysing the quality of the NBMs trained for the gear-bearing temperatures of the twelve WTs from the three wind farms. We assess the average of the NBMs for the individual WTs when trained with the respective learning strategy and training data time range. 
The accuracies of gear-bearing temperature NBMs trained with different learning strategies were averaged across all time
ranges, start dates, and all twelve wind turbines. As shown in Table \ref{tab:collaborative-learning}, intra-farm \FedAvg \ with fine-tuning resulted in NBMs with the lowest mean absolute errors (MAEs), followed by inter-farm learning with fine-tuning. Thus, NBMs trained collaboratively by multiple wind turbines can outperform those trained with only local data, especially when little training data (less than a few months, as in this study) is available. Furthermore, our results show that fine-tuning retains some of the collaborative knowledge, even though non-fine-tuned models exhibit moderate to poor performances, highlighting the importance of fine-tuning in consolidating collaborative learning gains and improving model performance. 

\begin{table}[H]
    \centering
    \renewcommand{\arraystretch}{1.25}
    \begin{tabular}{c|c|c|c|}
\cmidrule{2-4}
 & \multicolumn{2}{c|}{FedAvg} &   \multirow[t]{2}{*}{Local} \\
 \cmidrule{1-3}
  \multicolumn{1}{|c|}{Fine-tuned} & Intra-farm & Inter-farm &\\
\hline
\multicolumn{1}{|c|}{no} & 1.31 & 6.92 &   \\ 
\multicolumn{1}{|c|}{yes} & \textbf{0.87} & 0.97 &  \multirow[t]{2}{*}{1.44}\\ 
\hline
\end{tabular}
    \caption{Mean absolute errors (in °C) of the gear-bearing temperature NBMs trained with different learning strategies, averaged across all time ranges, start dates, and wind turbines.}
    \label{tab:collaborative-learning}
\end{table}

\begin{figure}[H]
    \centering
    \includegraphics[width=0.6\textwidth]{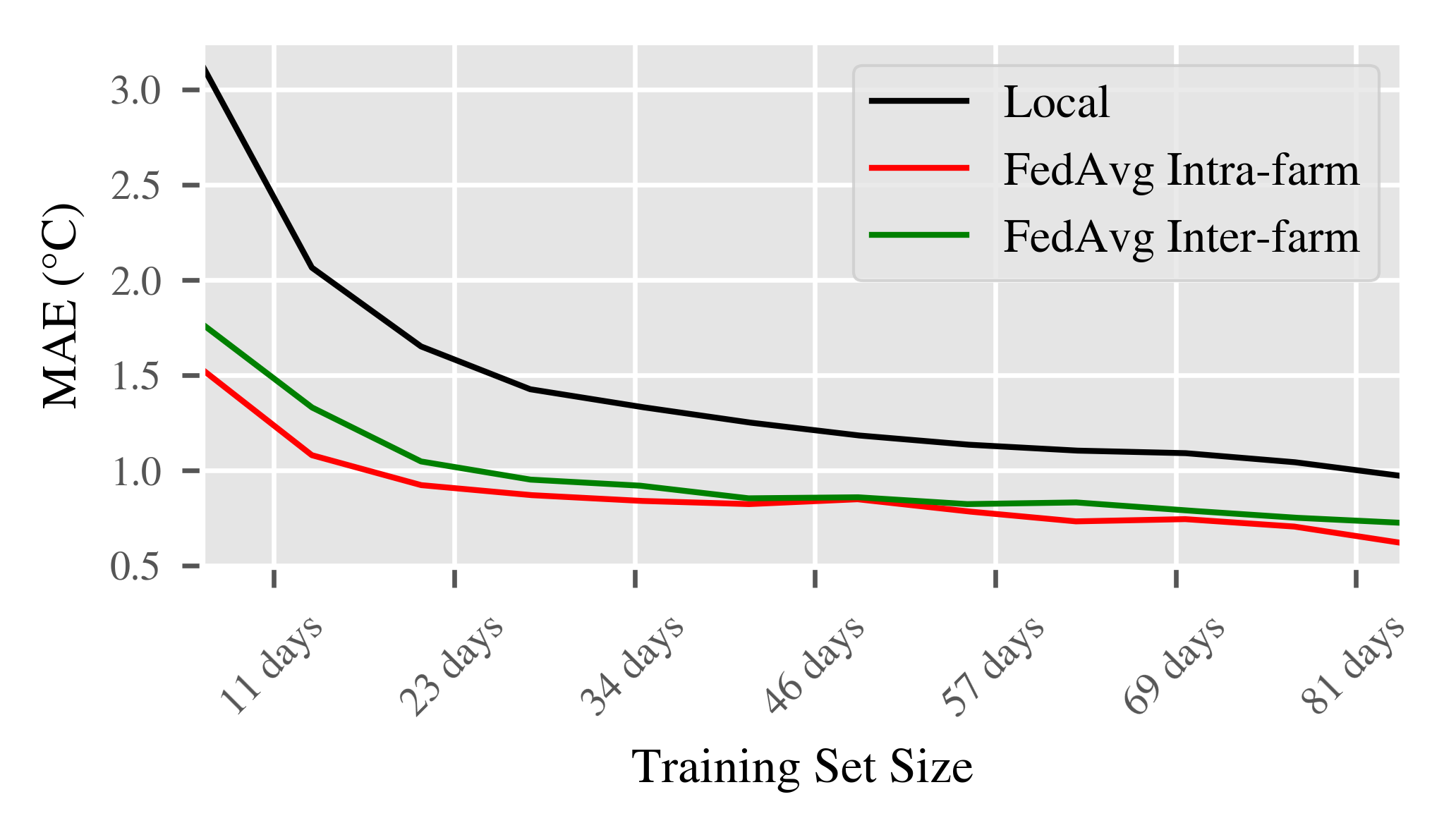}
    \caption{Mean absolute errors for fine-tuned \FedAvg \ and local training. The losses are averaged across all start dates and wind turbines.}
    \label{fig:time-evolution}
\end{figure}

Overall, intra-farm \FedAvg \ with fine-tuning demonstrates a significant improvement over local training, reducing the MAE by approximately 40\%. While inter-farm \FedAvg \ with fine-tuning also demonstrates improvement over local training, it falls slightly behind fine-tuned intra-farm learning. This discrepancy may be caused by the significant statistical heterogeneity between the wind farm SCADA variables involved in the NBM training. Intra-farm FL without fine-tuning outperforms local training by approximately 9\%. For inter-farm learning in presence of significant statistical heterogeneity across wind farms, the loss increases by almost fivefold without fine-tuning (Table \ref{tab:collaborative-learning}).
The relative performance between local training and fine-tuned intra- and inter-farm \FedAvg \ remains consistent across all time ranges considered, as shown in Figure \ref{fig:time-evolution}. FL consistently exhibits a strong improvement compared to local training, even with 12 weeks of training data. Our above results were averaged over different start dates. We investigated whether the results depend on the time of year and found they remain largely consistent across different seasons, as shown in Figure \ref{fig:seasonality} in the appendix. In all seasons, intra- and interfarm FL enable more accurate NBMs than local training. 
We also found our conclusions do not depend on the start date, with the exception of June 2017 (Figures \ref{fig:seasonality} and \ref{fig:output_plot_june_3w} in the appendix), which could possibly be due to a seasonality shift between the training and test set for the Kelmarsh wind farm.

\subsection{FL strongly reduces the training data collection time} 
\label{sec:cold-start-speed-up}
\begin{figure}[H]
    \centering
    \includegraphics[width=0.9\textwidth]{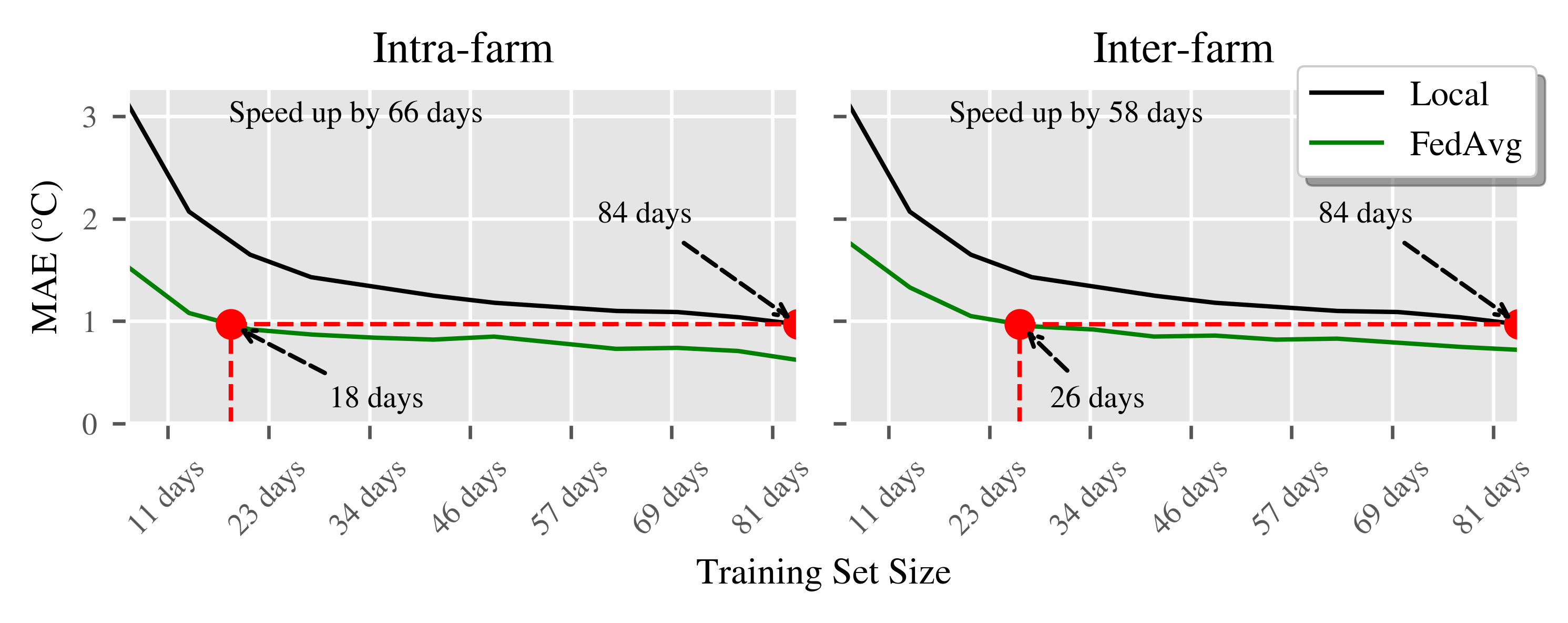}
    \caption{\FedAvg \ with fine-tuning reduces the time needed to accumulate the amount of historical training data required to achieve performances comparable to local training by 66 days in intra-farm learning and by 58 days in inter-farm learning, when averaged across all WTs and start dates.}
    \label{fig:fedavg-cold-start-speedup}
\end{figure}

The \emph{cold start speed up} refers to the reduction of time needed to accumulate the amount of historical training data required to achieve performances comparable to local training when employing FL. As shown in Figure \ref{fig:fedavg-cold-start-speedup}, \FedAvg \ with fine-tuning reduces the training data collection time by approximately eight to nine weeks out of the twelve weeks under consideration in intra-farm and inter-farm FL. For example in Figure \ref{fig:fedavg-cold-start-speedup}, to achieve the equivalent accuracy of the best performing gear bearing temperature NBM trained with local data, intra-farm \FedAvg \ with fine-tuning requires only 18 days of training data compared to the 84 days of training data required using local data only. This speed up allows for the earlier deployment of accurate NBMs, enabling an earlier fault detection and thereby reducing the risk of undetected incipient faults.

\subsection{Results by wind farm}
\label{sec:results-per-farm}
The statistical heterogeneity of the datasets of different clients can present a significant challenge to collaborative learning. Our results suggest that increasing the number of WTs involved in the training does not necessarily lead to improved collaborative learning outcomes, even after fine tuning. In particular, if the data distributions vary among the different wind turbines, the performance of collaborative learning across WTs of different wind farms (inter-farm learning) can be worse than that of collaborative learning within a given wind farm (intra-farm learning).

\begin{figure}[H]
    \centering
    \includegraphics[width=\textwidth]{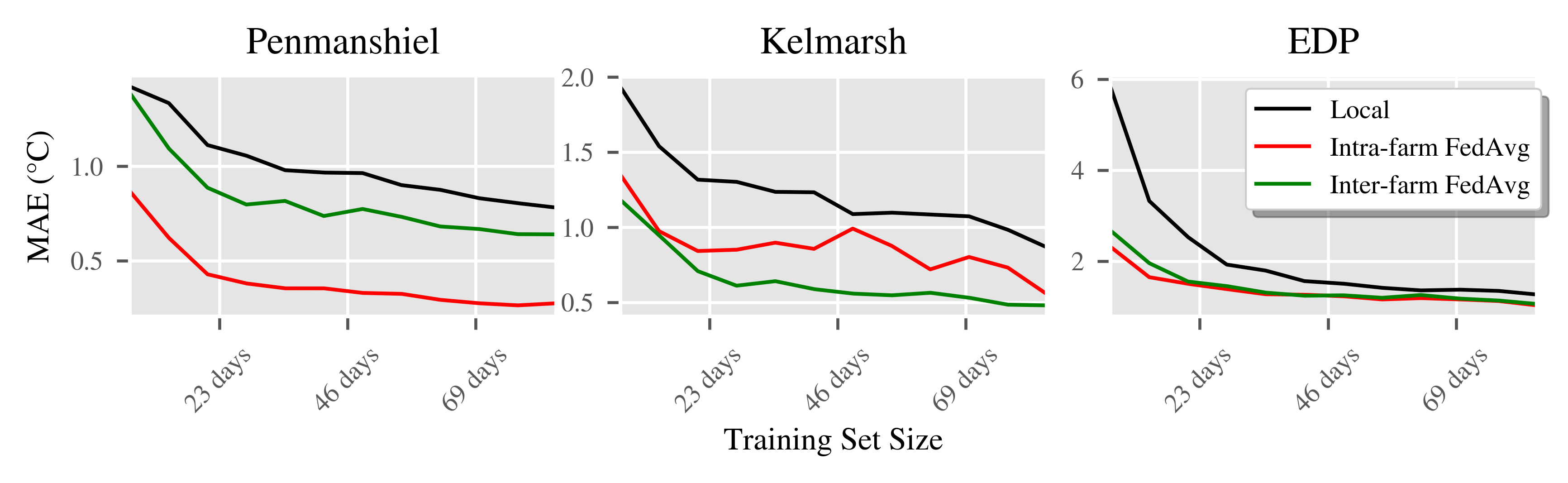}
    \caption{Evolution of the mean absolute errors (in °C) for fine-tuned \FedAvg \  and local training. These losses are the averaged results across all start dates and turbines within each wind farm.}
    \label{fig:farm-wise-time-evolution}
\end{figure}

We also assessed the performance of FL for NBM training in the context of the individual wind farms. We averaged the accuracies of the NBMs across the four turbines involved in the FL training from each wind farm, as shown in Table \ref{tab:farm-wise-collaborative-learning}. \FedAvg \ with fine-tuning consistently outperformed the other learning methods at each wind farm. Intra-farm FL with fine-tuning significantly surpasses the accuracy of inter-farm learning with fine-tuning at the Penmanshiel wind farm, as shown in Table \ref{tab:farm-wise-collaborative-learning} and Figure \ref{fig:farm-wise-time-evolution}. 
The comparatively poor performance of inter-farm FL is likely related to the low fraction of SCADA data from the Penmanshiel wind farm (see Figure \ref{fig:dataset_quantity} in the appendix), as its small data contribution to the inter-farm FL training results in a small contribution to the inter-farm FL model. 
The non-fine-tuned inter-farm MAE is substantially higher (16.65°C, see also Figure \ref{fig:output_plot_december_3w} in the appendix), indicating that the NBM primarily learns from the other two wind farms in the case of inter-farm learning.
Conversely, the EDP wind farm exhibits minimal disparity between intra- and inter-farm learning. This is likely because a significant portion of the global model's influence stems from EDP's data, which accounts for roughly $50\%$ of the total data across the various wind farms (as depicted in Figure \ref{fig:dataset_quantity}). Thus, the global model's performances on EDP's wind turbines are less affected by the heterogeneity introduced by other wind farms' data. 

\begin{table}[h]
    \centering
 \resizebox{0.4\columnwidth}{!}{\begin{tabular}{|l|c|c|c|c|}
\toprule
 &  & \multicolumn{2}{c|}{FedAvg} &  \multirow[t]{2}{*}{Local} \\
 & Fine-tuned  & Intra-farm & inter-farm &  \\
\midrule
\multirow[t]{2}{*}{EDP} & no & 1.74 & 2.33 &   \\
 & yes & \textbf{1.35} & 1.44 &  \multirow[t]{2}{*}{2.10}\\
\cline{1-5}
\noalign{\vskip-2\tabcolsep \vskip-3\arrayrulewidth \vskip\doublerulesep}
\\ \cline{1-5}
\multirow[t]{2}{*}{Kelmarsh} & no & 1.18 & 1.81 &  \\
 & yes & 0.87 & \textbf{0.65} & \multirow[t]{2}{*}{1.23}\\
\cline{1-5}
\noalign{\vskip-2\tabcolsep \vskip-3\arrayrulewidth \vskip\doublerulesep}
\\ \cline{1-5}
\multirow[t]{2}{*}{Penmanshiel} & no & 1.02 & 16.65 &   \\
 & yes & \textbf{0.40} & 0.82 &  \multirow[t]{2}{*}{1.00}\\
\cline{1-5}
\bottomrule
\end{tabular}}
    \caption{Mean absolute errors (in °C) of the gear-bearing temperature NBMs trained with different learning strategies from each wind farm, averaged across all time ranges, start dates, and wind turbines of the respective wind farm.}
    \label{tab:farm-wise-collaborative-learning}
\end{table}

\begin{figure}[H]
    \centering
    \includegraphics[width=\textwidth]{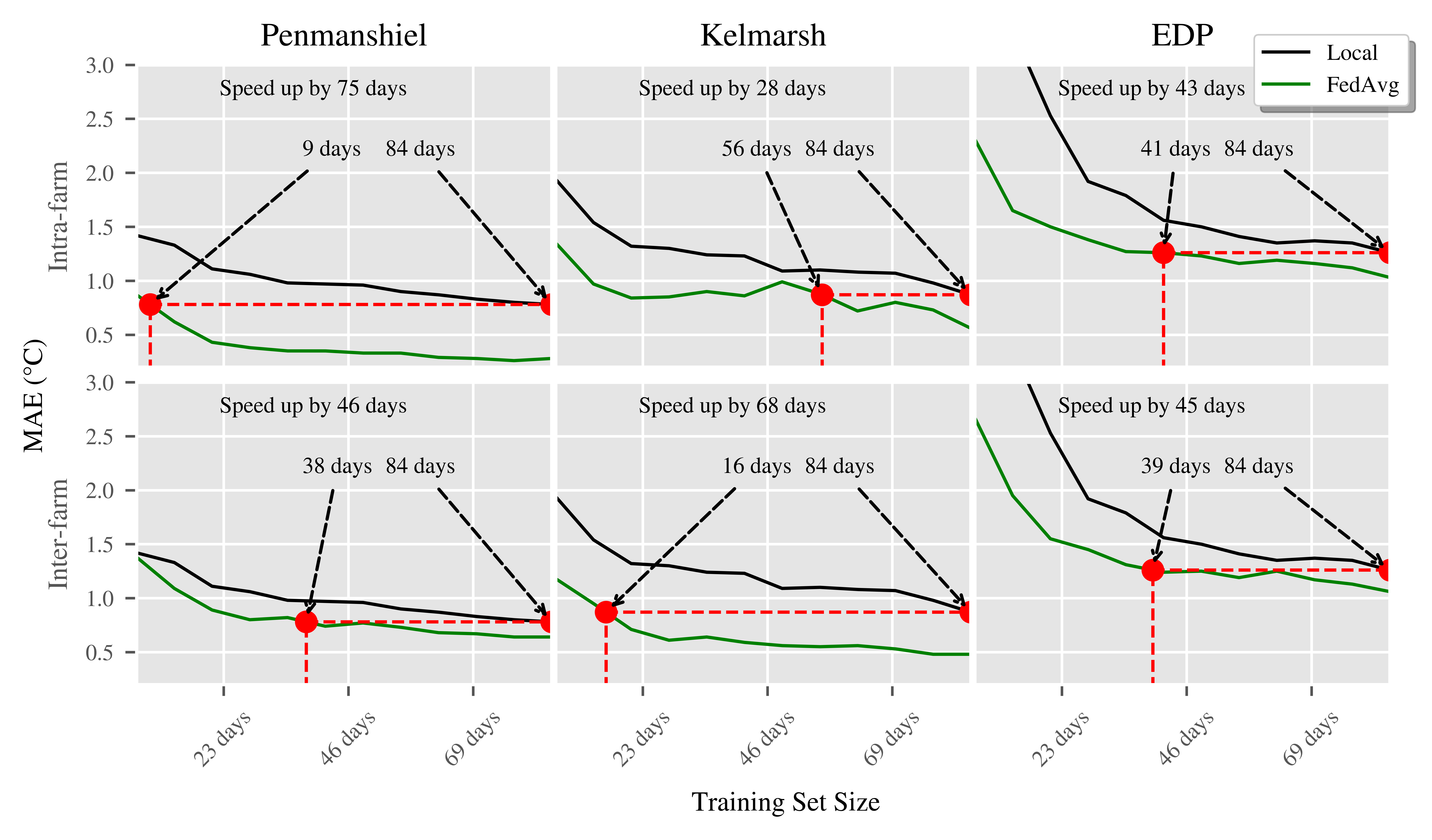}
    \caption{Cold start speed up for fine-tuned \FedAvg. These MAEs are averaged across the different start dates and turbines within each wind farm.}
    \label{fig:farm-wise-fedavg-cold-start-speedup}
\end{figure}

For each wind farm, the implementation of FL strategies results in a significant cold start speed up, with saved time ranging from four to more than ten weeks. This phenomenon is illustrated in Figure \ref{fig:farm-wise-fedavg-cold-start-speedup}, where the cold start speed up for fine-tuned \FedAvg \ is depicted. The most substantial reduction in the time required to collect training data occurs with intra-farm learning with fine-tuning in the Penmanshiel wind farm.

We considered four different start dates, one in each season, to investigate how seasonality impacts the accuracy of NBMs trained with different FL strategies at each wind farm. This resulted in twelve combinations of start date and wind farm, as shown in Figure \ref{fig:farm-seasonality} and Table \ref{tab:farm-wise-seasonal-collaborative-learning} in the appendix.
Intra-farm learning with fine-tuning emerged as the best-performing strategy in most cases (10 out of 12), and inter-farm learning with fine-tuning in the remaining two cases which pertain to the Kelmarsh wind farm.
We also investigated how the different learning strategies perform for individual wind turbines and found that our above results are confirmed also at WT level. Federated learning enables more accurate normal behaviour models than local training with limited data, and a reduction in the amount of time needed to collect the required model training data. Fine-tuning provided more accurate NBMs in all cases. Moreover, intra-farm FL tended to provide more accurate NBMs than inter-farm FL.

In addition to \FedAvg, all experiments were repeated with an alternative federated learning algorithm called \FedProx \ \cite{FederatedOptimizationinHeterogeneousNetworks}. \FedProx \ follows a similar learning process as \FedAvg \ but incorporates a regularization term in the loss function during local training, which measures the discrepancy between the current global model and the updated local model of the clients. Our experiment indicates that \FedProx \ performs comparably, albeit slightly worse, than \FedAvg \ for both inter-farm and intra-farm learning. This may be attributed to the fact that \FedProx \ 
slows down local training to retain more knowledge from the global FL model but does not address the models competing against each other due to statistical heterogeneity. 

\begin{figure}[H]
    \centering
    \includegraphics[width=\textwidth]{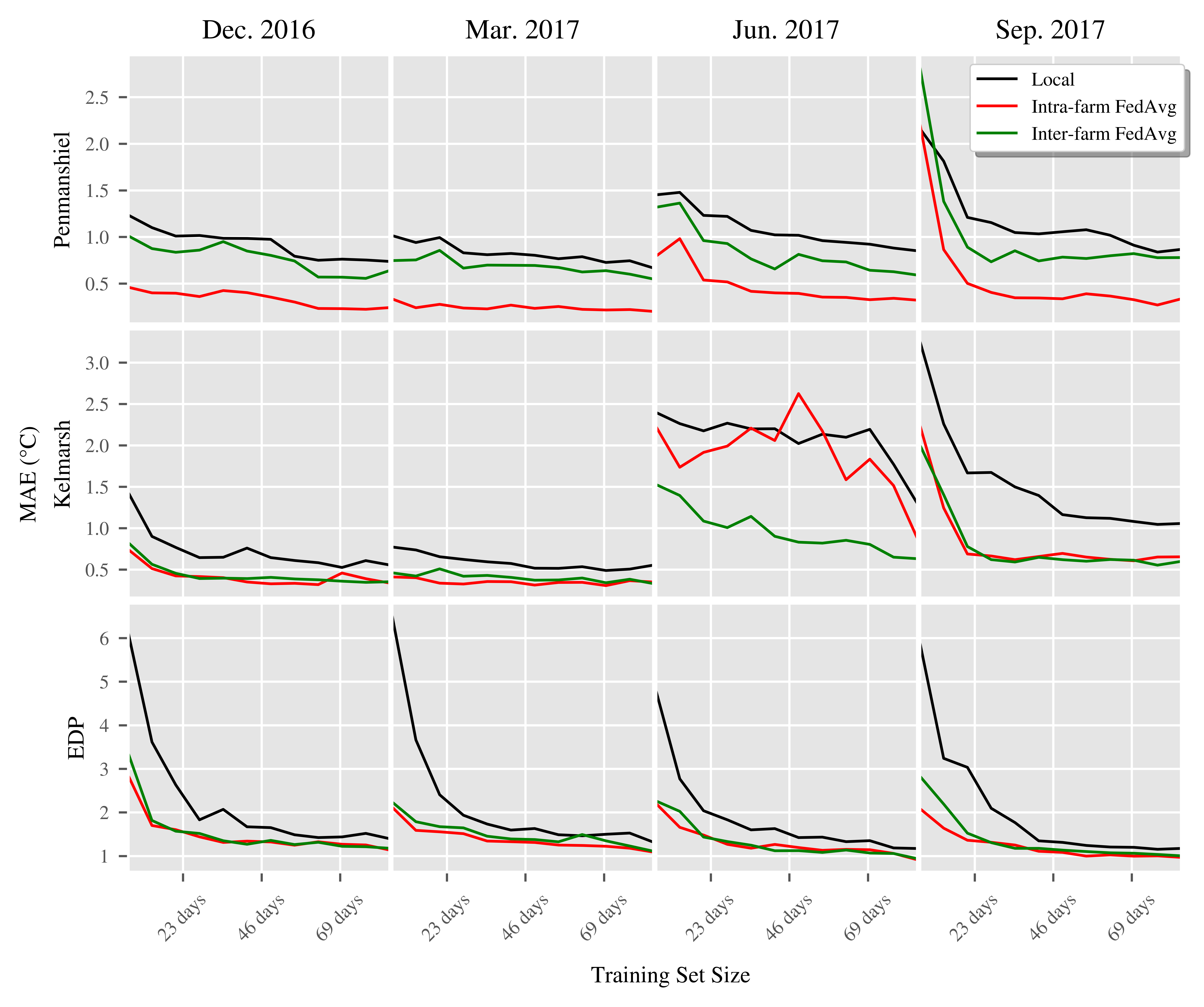}
    \caption{Mean absolute error for each start date and wind farm. These results are averaged across the turbines within a given wind farm.}
    \label{fig:farm-seasonality}
\end{figure}

\newpage
\section{Conclusions}

Our study demonstrated the effectiveness of privacy-preserving collaborative learning of wind turbine normal behaviour models for condition monitoring applications. We demonstrated that federated learning enables a reduction of the training data accumulation time, allowing for an earlier detection of developing faults, compared to local training. We also found that having more collaborators is not necessarily better in collaborative learning. In the presence of high statistical heterogeneity, i.e., significant differences between the data distributions of the involved wind turbines, the performance of federated learning across wind turbines of different wind farms (\emph{inter-farm} learning) is worse than that of federated learning within a given wind farm (\emph{intra-farm} learning). 

We assessed two distinct collaborative learning approaches: inter-farm learning, which involves collaboration across turbines from different wind farms, and intra-farm learning, which restricts collaboration to turbines in the same wind farm. Our analysis shows that high levels of statistical heterogeneity present significant challenges to collaborative learning. The accuracy of NBMs trained in intra-farm learning surpassed that of inter-farm learning in most situations, underscoring the adverse impacts of heterogeneity on collaborative learning. We demonstrated fine-tuning as an approach to address FL model training in view of significant statistical heterogeneity.

We propose several directions of future research. Firstly, our model selection and hyperparameter tuning processes were conducted on a full dataset from a single turbine, potentially diverging from real-world conditions where historical data accumulation occurs incrementally and disregarding the contributions of the other turbines. Addressing this challenge entails the development of automated and adaptive model selection methods capable of accommodating evolving data volumes and complexities \cite{AdaptationStrategiesforAutomatedMachineLearningonEvolvingData}. Moreover, integrating FL techniques for hyperparameter tuning \cite{FederatedHyperparameterTuning:ChallengesBaselinesandConnectionstoWeight-Sharing} could enhance efficiency and scalability in FL settings. Furthermore, we propose to investigate continuous learning strategies \cite{AComprehensiveSurveyofContinualLearning:TheoryMethodandApplication} for their potential to reduce communication costs and training time in FL by enabling incremental model updates instead of periodic full re-training. Incorporating such strategies would not only contribute to the ongoing evolution and refinement of collaborative learning methodologies in renewable energy applications but may also render them more applicable to real industrial settings. Finally, we considered identical feature and label spaces of all client wind turbines but did not consider other types of FL in this study, such as federated transfer learning \cite{FederatedTransferLearning}. They may form subject of future research in wind energy applications.

In conclusion, our study demonstrated the potential and challenges of collaborative learning in wind turbine condition monitoring through FL. By advancing our understanding of effective collaboration strategies and addressing challenges such as statistical heterogeneity and model adaptation, we move closer to realizing the full potential of FL for enhancing the reliability and efficiency of wind farms and other renewable energy systems.

\section*{CRediT author statement}

\textbf{Grataloup, A.}: Conceptualization, Formal analysis, Investigation, Methodology, Software, Validation, Visualization, Writing -- original draft, Writing -- review \& editing. 

\textbf{Jonas, S.}: Conceptualization, Methodology, Writing -- review \& editing.

\textbf{Meyer, A.}: Conceptualization, Writing -- review \& editing, Funding acquisition, Project administration, Supervision, Resources.
\section*{Acknowledgment} 

This research was supported by the Swiss National Science Foundation (Grant No. 206342).

\newpage
\section*{Appendix}
\appendix

\section{}

\setcounter{figure}{0}
\setcounter{table}{0}

\begin{figure}[H]
    \centering
    \includegraphics{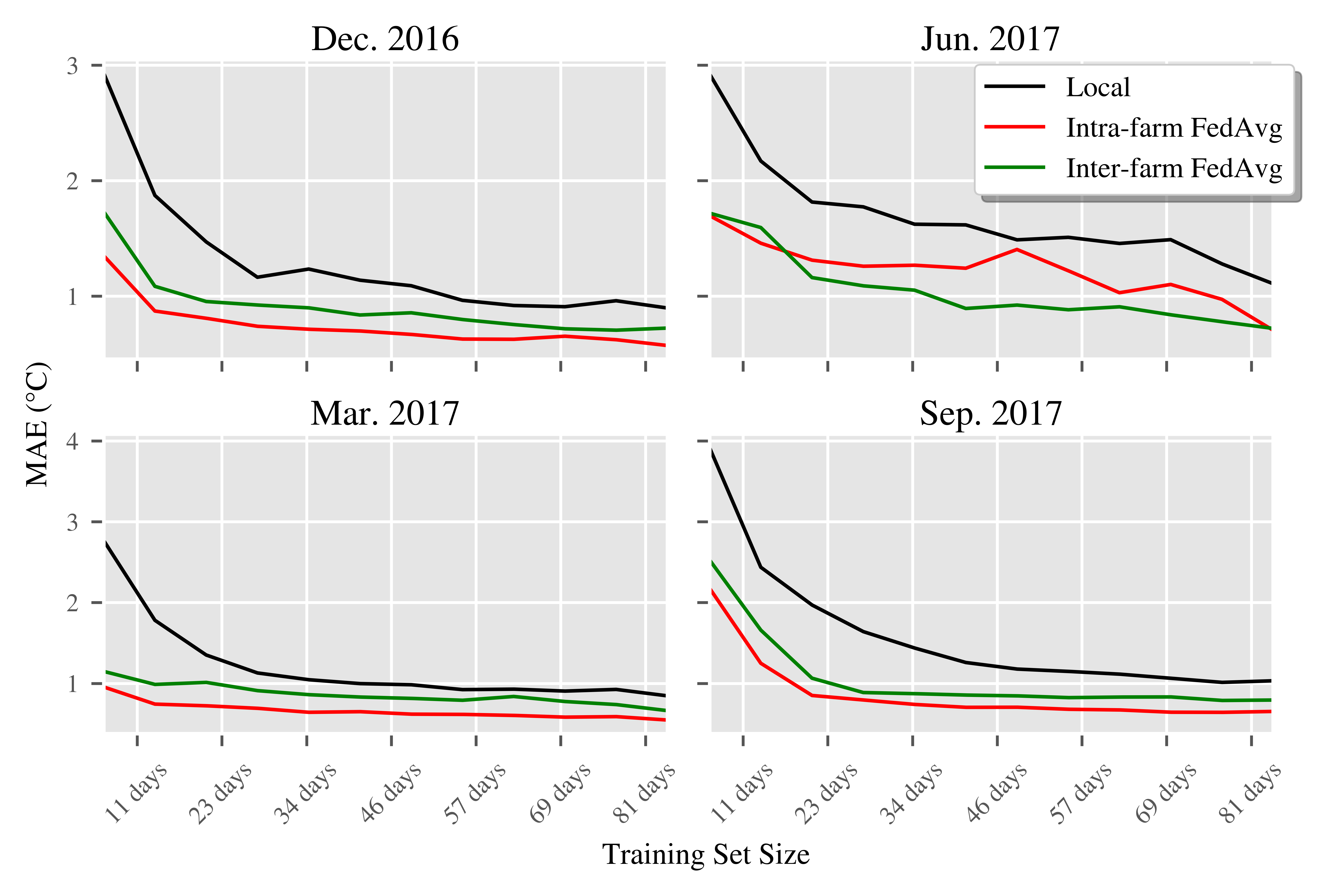}
    \caption{Mean absolute errors for each start date, with fine-tuned \FedAvg \ and local training. These results are averaged across all turbines.}
    \label{fig:seasonality}
\end{figure}

Figure \ref{fig:seasonality} shows that the evolution of the mean absolute errors for fine-tuned \texttt{FedAvg} and local training by starting date is also consistent across all seasons with the exception of the starting date in June 2017. A possible explanation for this behavior is a seasonality-based data distribution shift between the training and test set for the Kelmarsh wind farm, which is discussed further in Appendix B.

\section{}
\setcounter{figure}{0}
\setcounter{table}{0}

\begin{figure}[H]
  \centering
 \includegraphics[width=1\textwidth]{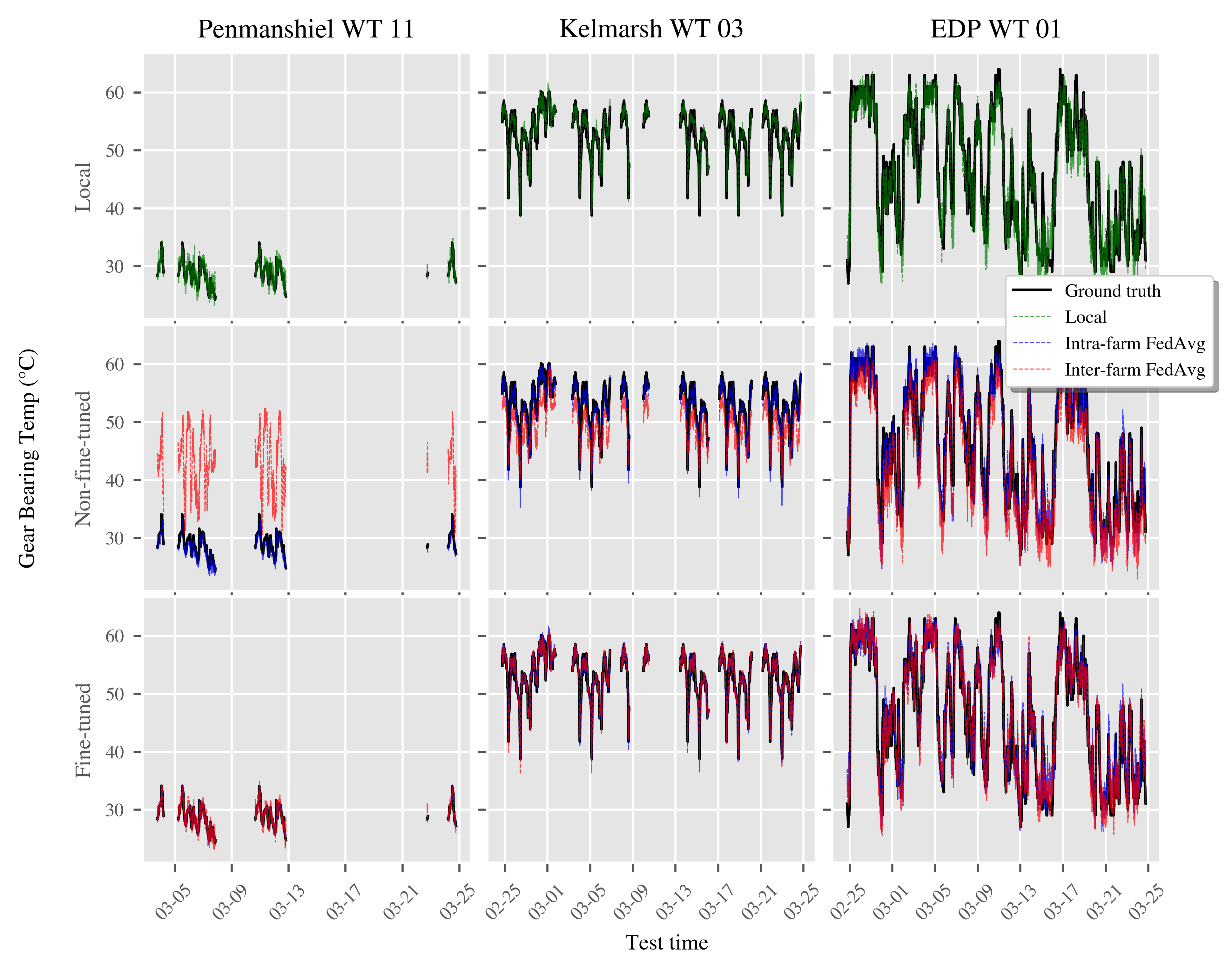}
  \caption{Actual versus predicted gear bearing temperatures using local learning (1\ts{st} row), intra and inter-farm \FedAvg \ with and without finetuning (3\ts{rd} and 2\ts{nd} row, respectively).}
  \label{fig:output_plot_december_3w}
\end{figure}

We examine the model prediction compared to the ground truth of the gear-bearing temperature for one selected turbine from each wind farm. We restrict ourselves here to models trained on three weeks of training data in December for Figure \ref{fig:output_plot_december_3w} and June for Figure \ref{fig:output_plot_june_3w}, using the corresponding four-week test dataset 
for evaluation. This analysis provides insights into the performance of various learning methods, enabling a qualitative assessment of FL. Notably, predictions from local training (no collaboration) and fine-tuned \FedAvg \ closely align with the ground truth. However, non-fine-tuned \FedAvg \ exhibits inferior performance, especially in scenarios characterized by significant data heterogeneity among clients, such as inter-farm learning.

The models were trained on three weeks of data in December 2016. Figure \ref{fig:output_plot_december_3w} illustrates the performance of the local, intra- and interfarm learning strategies on the respective test sets. A single WT has been picked randomly from each wind farm to this end. 

One important observation from Figure \ref{fig:output_plot_december_3w} is the widening error of non-fine-tuned inter-farm \FedAvg \ as the data proportion decreases. Specifically, the red curve representing non-fine-tuned inter-farm \FedAvg \ for Penmanshiel (the wind farm with the smallest data proportion) shows a pronounced deviation from ground truth and is closer to the temperature ranges observed in the Kelmarsh and EDP wind farms. This disparity arises from the observable heterogeneity, with Penmanshiel exhibiting a temperature range of 25°C to 35°C, contrasting with the 40°C to 60°C range observed in Kelmarsh and EDP wind farms. \FedAvg \ attributes a larger weight to clients with more data, resulting in Penmanshiel's turbine contributing significantly less than those of Kelmarsh and EDP wind farms (see Figure \ref{fig:dataset_quantity}).

Non-fine-tuned intra-farm learning outperforms non-fine-tuned inter-farm learning in this case study. This indicates that the different wind farms involved in the collaborative FL process end up competing to achieve different learning objectives rather than collaborate. However, this disparity is no longer visible after fine-tuning and opens the question of whether there is a retention of collaborative knowledge. 

While these observations hold across most start dates and time ranges, a different behaviour emerges in situations where the local training fails to fit the test set, as shown in Figure \ref{fig:output_plot_june_3w}:

\begin{figure}[H]
  \centering
  \includegraphics[width=1\textwidth]{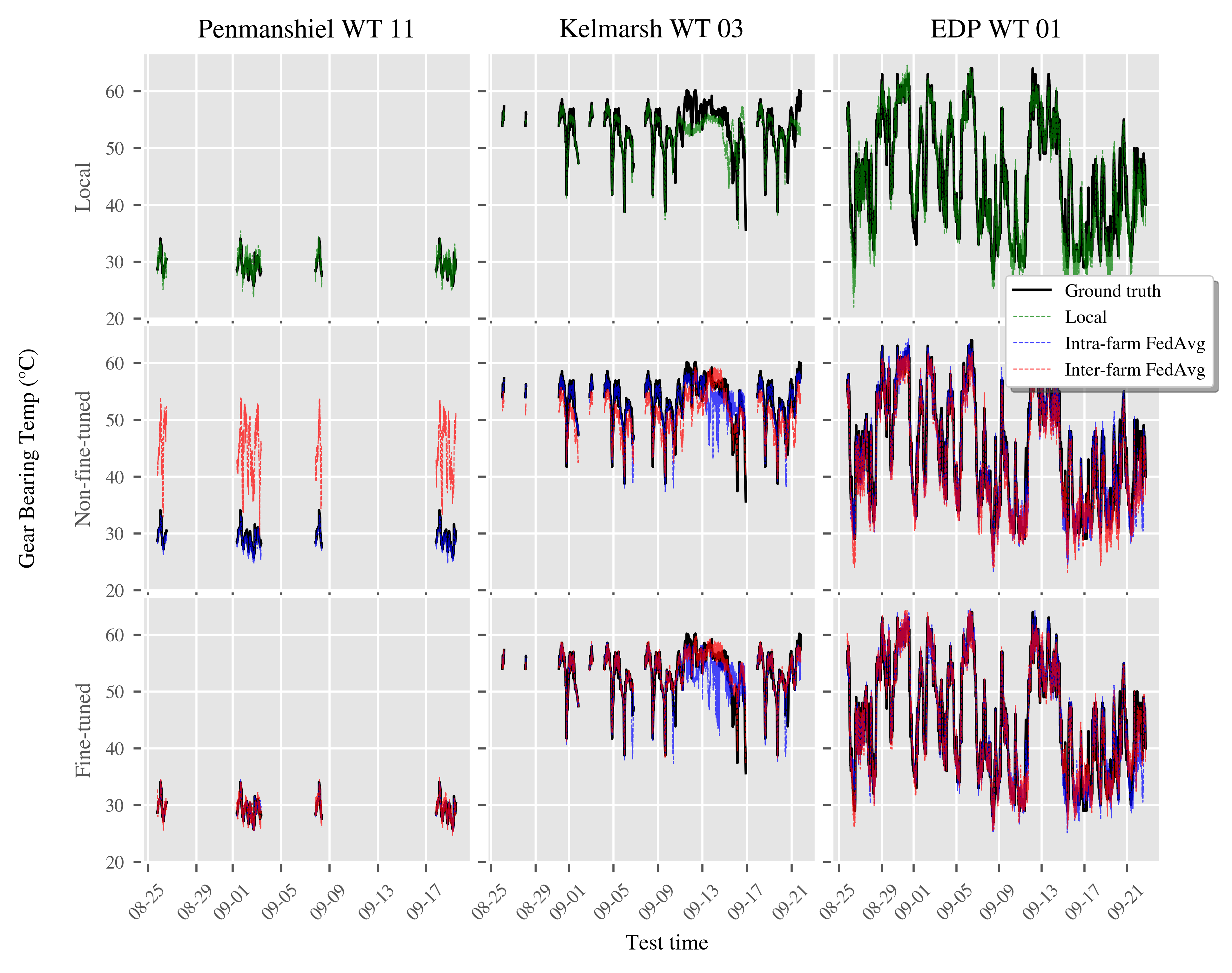}
  \caption{For each wind farm, we compare ground truth (in black) to the predicted gear bearing temperature (in °C) using local learning (1\ts{st} row), non-fine-tuned intra and inter-farm \FedAvg \ (2\ts{nd} row), and their fine-tuned version (3\ts{rd} row). A single turbine has been picked for each wind farm. This visual evaluation is done on the test dataset and the models are trained on three weeks of data in June 2017.}
  \label{fig:output_plot_june_3w}
\end{figure}

In Figure \ref{fig:output_plot_june_3w}, the overall observations remain consistent, except for the Kelmarsh wind farm during the testing period between September 11 and September 17. During this period, local training fails to effectively fit the ground truth, likely indicating either potential anomalies in the data, or data points lying outside the range of the model’s training data, leading to poor generalization. Upon examining the data distribution of various features, we find no indications of anomalous behavior during that
period. Furthermore, we observe wind speed and ambient temperature value ranges slightly higher and lower, respectively, in the affected test set compared to the training set. Such variations are expected when comparing weather conditions between June and September. A distribution shift being a possible cause is further supported by the observation that during this period, inter-farm learning, both with and without fine-tuning, outperforms intra-farm learning (see also Figure \ref{fig:seasonality}). This suggests that inter-farm learning may benefit from insights gained from other farms, enabling it to better adapt to locally unseen data. However, in practical scenarios, our primary concern is not whether a model trained on data from June will perform well on test data in September. Instead, our focus would lie on ensuring that the model performs effectively in the weeks that follow the training set. We could continuously retrain the model with new data as it becomes available, thereby mitigating any seasonality shift and increasing the likelihood of the model performing well on near-future data in a continuous learning setting.

\section{} 
\setcounter{figure}{0}
\setcounter{table}{0}

\begin{figure}[H]
  \centering
  \includegraphics{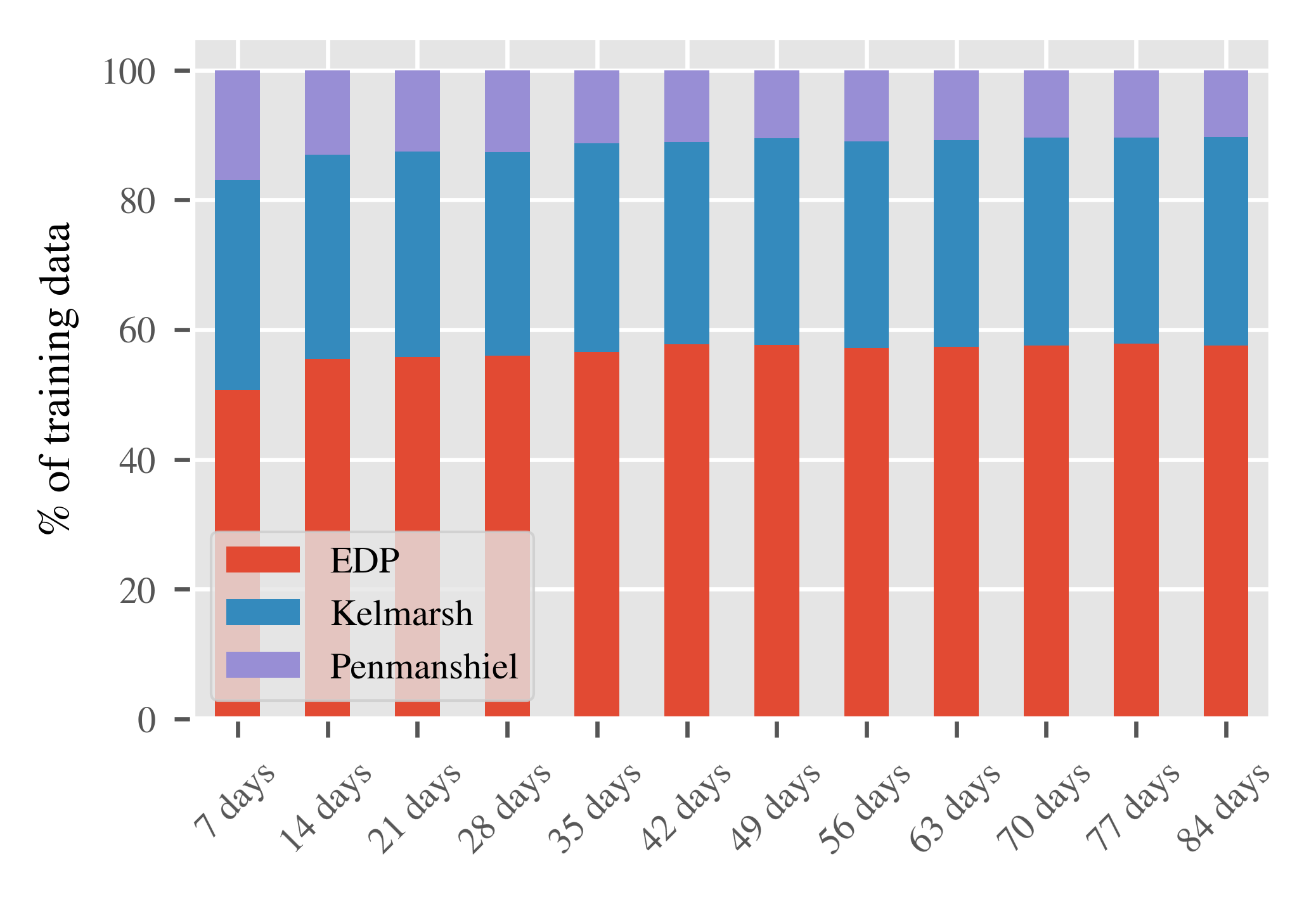}
  \caption{Proportion of data available per wind farm when taking different time windows of data (1 to 12 weeks by 1-week intervals).}
  \label{fig:dataset_quantity}
\end{figure}

\begin{table}[h]
    \centering
    \resizebox{0.85\columnwidth}{!}{\begin{tabular}{|l|c||c|c|c||c|c|c||c|c|c|}
\toprule
 &  & \multicolumn{3}{c||}{EDP} & \multicolumn{3}{c||}{Kelmarsh} & \multicolumn{3}{c|}{Penmanshiel} \\
 \cmidrule{3-11}
 &  & \multicolumn{2}{c|}{FedAvg} & \multirow[t]{2}{*}{Local} & \multicolumn{2}{c|}{FedAvg} & \multirow[t]{2}{*}{Local} & \multicolumn{2}{c|}{FedAvg} &  \multirow[t]{2}{*}{Local} \\
 & Fine-tuned & Intra-farm & Inter-farm &  & Intra-farm & Inter-farm &  & Intra-farm & Inter-farm &  \\
\midrule
\multirow[t]{2}{*}{2016-12-01} & no & 1.79 & 2.49 &  & 0.81 & 1.82 &  & 0.95 & 17.80 &  \\
 & yes & \textbf{1.48} & 1.53 & \multirow[t]{2}{*}{2.24} & \textbf{0.42 }& 0.44 & \multirow[t]{2}{*}{0.72} & \textbf{0.34} & 0.77 & \multirow[t]{2}{*}{0.93} \\
\cline{1-11}
\noalign{\vskip-2\tabcolsep \vskip-3\arrayrulewidth \vskip\doublerulesep}
\\ \cline{1-11}
\multirow[t]{2}{*}{2017-03-01} & no & 1.67 & 2.47 &  & 0.78 & 1.93 &  & 0.94 & 16.32 &  \\
 & yes & \textbf{1.39} & 1.51 & \multirow[t]{2}{*}{2.23} & \textbf{0.35} & 0.40 & \multirow[t]{2}{*}{0.59} & \textbf{0.25} & 0.68 & \multirow[t]{2}{*}{0.83} \\
\cline{1-11}
\noalign{\vskip-2\tabcolsep \vskip-3\arrayrulewidth \vskip\doublerulesep}
\\ \cline{1-11}
\multirow[t]{2}{*}{2017-06-01} & no & 1.77 & 2.02 &  & 2.08 & 1.85 &  & 1.06 & 13.76 &  \\
 & yes & \textbf{1.30} & 1.32 & \multirow[t]{2}{*}{1.88} & 1.89 & \textbf{0.96} & \multirow[t]{2}{*}{2.08} & \textbf{0.48} & 0.85 &\multirow[t]{2}{*}{1.09} \\
\cline{1-11}
\noalign{\vskip-2\tabcolsep \vskip-3\arrayrulewidth \vskip\doublerulesep}
\\ \cline{1-11}
\multirow[t]{2}{*}{2017-09-01} & no & 1.74 & 2.34 &  & 1.06 & 1.63 &  & 1.12 & 18.76 &  \\
 & yes & \textbf{1.24} & 1.38 & \multirow[t]{2}{*}{2.05} & 0.83 & \textbf{0.80} & \multirow[t]{2}{*}{1.53} & \textbf{0.52} & 0.97 & \multirow[t]{2}{*}{1.16} \\
\cline{1-11}
\bottomrule
\end{tabular}}
    \caption{Mean absolute error (in °C) for each wind farm, start date and learning strategy, averaged across all time ranges, and wind turbines within each wind farm.}
    \label{tab:farm-wise-seasonal-collaborative-learning}
\end{table}

\bibliographystyle{elsarticle-num}
\bibliography{main}

\begin{thebibliography}{10}
\expandafter\ifx\csname url\endcsname\relax
  \def\url#1{\texttt{#1}}\fi
\expandafter\ifx\csname urlprefix\endcsname\relax\def\urlprefix{URL }\fi
\expandafter\ifx\csname href\endcsname\relax
  \def\href#1#2{#2} \def\path#1{#1}\fi

\bibitem{euwindreoprt2023}
T.~E, T.~J, S.~A, G.~A, D.~M, L.~S, K.~A, M.~A, I.~E, S.~D, J.~O. G, E.~OD, G.~M, Clean energy technology observatory: Wind energy in the european union - 2023 status report on technology development, trends, value chains and markets~(KJ-NA-31-678-EN-N (online)).
\newblock \href {http://dx.doi.org/10.2760/618644 (online)} {\path{doi:10.2760/618644 (online)}}.

\bibitem{globalwindreport2023}
GWEC, Global wind report 2023, \url{https://gwec.net/wp-content/uploads/2023/03/GWR-2023_interactive.pdf}.

\bibitem{DeepLearningforAnomalyDetection:AReview}
G.~Pang, C.~Shen, L.~Cao, A.~V.~D. Hengel, \href{https://doi.org/10.1145/3439950}{Deep learning for anomaly detection: A review}, ACM Comput. Surv. 54~(2).
\newblock \href {http://dx.doi.org/10.1145/3439950} {\path{doi:10.1145/3439950}}.
\newline\urlprefix\url{https://doi.org/10.1145/3439950}

\bibitem{AnomalyDetectiononWindTurbinesBasedonaDeepLearningAnalysisofVibrationSignals}
J.~L.~C. Hoffmann, L.~P. Horstmann, M.~M. Lucena, G.~M. de~Araujo, A.~A. Fröhlich, M.~H.~N. Nishioka, \href{https://doi.org/10.1080/08839514.2021.1966879}{Anomaly detection on wind turbines based on a deep learning analysis of vibration signals}, Applied Artificial Intelligence 35~(12) (2021) 893--913.
\newblock \href {http://arxiv.org/abs/https://doi.org/10.1080/08839514.2021.1966879} {\path{arXiv:https://doi.org/10.1080/08839514.2021.1966879}}, \href {http://dx.doi.org/10.1080/08839514.2021.1966879} {\path{doi:10.1080/08839514.2021.1966879}}.
\newline\urlprefix\url{https://doi.org/10.1080/08839514.2021.1966879}

\bibitem{AnomalyDetectionforWindTurbinesUsingLongShort-TermMemory-BasedVariationalAutoencoderWassersteinGenerationAdversarialNetworkunderSemiSupervisedTraining}
C.~Zhang, T.~Yang, \href{https://www.mdpi.com/1996-1073/16/19/7008}{Anomaly detection for wind turbines using long short-term memory-based variational autoencoder wasserstein generation adversarial network under semi-supervised training}, Energies 16~(19).
\newblock \href {http://dx.doi.org/10.3390/en16197008} {\path{doi:10.3390/en16197008}}.
\newline\urlprefix\url{https://www.mdpi.com/1996-1073/16/19/7008}

\bibitem{AnomalyDetectiononSmallWindTurbineBladesUsingDeepLearningAlgorithms}
B.~Altice, E.~Nazario, M.~Davis, M.~Shekaramiz, T.~K. Moon, M.~A.~S. Masoum, \href{https://www.mdpi.com/1996-1073/17/5/982}{Anomaly detection on small wind turbine blades using deep learning algorithms}, Energies 17~(5).
\newblock \href {http://dx.doi.org/10.3390/en17050982} {\path{doi:10.3390/en17050982}}.
\newline\urlprefix\url{https://www.mdpi.com/1996-1073/17/5/982}

\bibitem{Maron2022}
J.~Maron, D.~Anagnostos, B.~Brodbeck, A.~Meyer, Artificial intelligence-based condition monitoring and predictive maintenance framework for wind turbines, Journal of Physics: Conference Series (2022) 2151\href {http://dx.doi.org/10.1088/1742-6596/2151/1/012007} {\path{doi:10.1088/1742-6596/2151/1/012007}}.

\bibitem{Jonas2023}
S.~Jonas, D.~Anagnostos, B.~Brodbeck, A.~Meyer, Vibration fault detection in wind turbines based on normal behaviour models without feature engineering, Energies (2023) 1760\href {http://dx.doi.org/10.3390/en16041760} {\path{doi:10.3390/en16041760}}.

\bibitem{Anartificialneuralnetwork-basedconditionmonitoringmethodforwindturbineswithapplicationtothemonitoringofthegearbox}
P.~Bangalore, S.~Letzgus, D.~Karlsson, M.~Patriksson, \href{https://onlinelibrary.wiley.com/doi/abs/10.1002/we.2102}{An artificial neural network-based condition monitoring method for wind turbines, with application to the monitoring of the gearbox}, Wind Energy 20~(8) (2017) 1421--1438.
\newblock \href {http://arxiv.org/abs/https://onlinelibrary.wiley.com/doi/pdf/10.1002/we.2102} {\path{arXiv:https://onlinelibrary.wiley.com/doi/pdf/10.1002/we.2102}}, \href {http://dx.doi.org/https://doi.org/10.1002/we.2102} {\path{doi:https://doi.org/10.1002/we.2102}}.
\newline\urlprefix\url{https://onlinelibrary.wiley.com/doi/abs/10.1002/we.2102}

\bibitem{RenewablesSharedataonwindenergy}
A.~Kusiak, \href{https://doi.org/10.1038/529019a}{Renewables: Share data on wind energy}, Nature 529~(7584) (2016) 19--21.
\newblock \href {http://dx.doi.org/10.1038/529019a} {\path{doi:10.1038/529019a}}.
\newline\urlprefix\url{https://doi.org/10.1038/529019a}

\bibitem{fl_review_grataloup_jonas_meyer}
A.~Grataloup, S.~Jonas, A.~Meyer, \href{https://www.sciencedirect.com/science/article/pii/S2666546824000417}{A review of federated learning in renewable energy applications: Potential, challenges, and future directions}, Energy and AI 17 (2024) 100375.
\newblock \href {http://dx.doi.org/https://doi.org/10.1016/j.egyai.2024.100375} {\path{doi:https://doi.org/10.1016/j.egyai.2024.100375}}.
\newline\urlprefix\url{https://www.sciencedirect.com/science/article/pii/S2666546824000417}

\bibitem{liReviewApplicationsFederated2020}
L.~Li, Y.~Fan, M.~Tse, K.-Y. Lin, A review of applications in federated learning, Computers \& Industrial Engineering 149 (2020) 106854.
\newblock \href {http://dx.doi.org/10.1016/j.cie.2020.106854} {\path{doi:10.1016/j.cie.2020.106854}}.

\bibitem{Windpowerforecastingconsideringdataprivacyprotection}
Y.~Li, R.~Wang, Y.~Li, M.~Zhang, C.~Long, Wind power forecasting considering data privacy protection: {{A}} federated deep reinforcement learning approach, Applied Energy 329~(C).
\newblock \href {http://dx.doi.org/10.1016/j.apenergy.2022.1} {\path{doi:10.1016/j.apenergy.2022.1}}.

\bibitem{DeepFederatedLearning-BasedPrivacy-PreservingWindPowerForecasting}
A.~Ahmadi, M.~Talaei, M.~Sadipour, A.~M. Amani, M.~Jalili, Deep federated learning-based privacy-preserving wind power forecasting, IEEE access : practical innovations, open solutions 11 (2023) 39521--39530.
\newblock \href {http://dx.doi.org/10.1109/ACCESS.2022.3232475} {\path{doi:10.1109/ACCESS.2022.3232475}}.

\bibitem{ACyber-Securegeneralizedsupermodelforwindpowerforecastingbasedondeepfederatedlearningandimageprocessing}
H.~Moayyed, A.~Moradzadeh, B.~{Mohammadi-Ivatloo}, A.~P. Aguiar, R.~Ghorbani, A {{Cyber-Secure}} generalized supermodel for wind power forecasting based on deep federated learning and image processing, Energy Conversion and Management 267 (2022) 115852.
\newblock \href {http://dx.doi.org/10.1016/j.enconman.2022.115852} {\path{doi:10.1016/j.enconman.2022.115852}}.

\bibitem{APrivacy-preservingWindSpeedPredictionMethodBasedonFederatedDeepLearning}
Y.~Wang, W.~Zhang, Q.~Guo, Y.~Wu, A privacy-preserving wind speed prediction method based on federated deep learning, in: 2022 4th International Conference on Power and Energy Technology ({{ICPET}}), 2022, pp. 638--643.
\newblock \href {http://dx.doi.org/10.1109/ICPET55165.2022.9918344} {\path{doi:10.1109/ICPET55165.2022.9918344}}.

\bibitem{AnefficientfederatedtransferlearningframeworkforcollaborativemonitoringofwindturbinesinIoE-enabledwindfarms}
L.~Wang, W.~Fan, G.~Jiang, P.~Xie, \href{https://www.sciencedirect.com/science/article/pii/S0360544223019126}{An efficient federated transfer learning framework for collaborative monitoring of wind turbines in {IoE}-enabled wind farms}, Energy 284 (2023) 128518.
\newblock \href {http://dx.doi.org/https://doi.org/10.1016/j.energy.2023.128518} {\path{doi:https://doi.org/10.1016/j.energy.2023.128518}}.
\newline\urlprefix\url{https://www.sciencedirect.com/science/article/pii/S0360544223019126}

\bibitem{Windturbinebladeicingdetection:afederatedlearningapproach}
X.~Cheng, F.~Shi, Y.~Liu, X.~Liu, L.~Huang, Wind turbine blade icing detection: A federated learning approach, Energy 254 (2022) 124441.
\newblock \href {http://dx.doi.org/10.1016/j.energy.2022.124441} {\path{doi:10.1016/j.energy.2022.124441}}.

\bibitem{AClass-ImbalancedHeterogeneousFederatedLearningModelforDetectingIcingonWindTurbineBlades}
X.~Cheng, F.~Shi, Y.~Liu, J.~Zhou, X.~Liu, L.~Huang, A class-imbalanced heterogeneous federated learning model for detecting icing on wind turbine blades, IEEE Transactions on Industrial Informatics 18~(12) (2022) 8487--8497.
\newblock \href {http://dx.doi.org/10.1109/TII.2022.3167467} {\path{doi:10.1109/TII.2022.3167467}}.

\bibitem{HumanKnowledge-basedCompressedFederatedLearningModelforWindTurbineBladeIcingDetection}
D.~Zhang, W.~Tian, Y.~Yin, X.~Liu, X.~Cheng, F.~Shi, Human knowledge-based compressed federated learning model for wind turbine blade icing detection, in: 2022 International Conference on High Performance Big Data and Intelligent Systems ({{HDIS}}), 2022, pp. 277--281.
\newblock \href {http://dx.doi.org/10.1109/HDIS56859.2022.9991642} {\path{doi:10.1109/HDIS56859.2022.9991642}}.

\bibitem{ABlockchain-EmpoweredCluster-BasedFederatedLearningModelforBladeIcingEstimationonIoT-EnabledWindTurbine}
X.~Cheng, W.~Tian, F.~Shi, M.~Zhao, S.~Chen, H.~Wang, A blockchain-empowered cluster-based federated learning model for blade icing estimation on {{IoT-Enabled}} wind turbine, IEEE Transactions on Industrial Informatics 18~(12) (2022) 9184--9195.
\newblock \href {http://dx.doi.org/10.1109/TII.2022.3159684} {\path{doi:10.1109/TII.2022.3159684}}.

\bibitem{DeepFedWT:Afederateddeeplearningframeworkforfaultdetectionofwindturbines}
G.~Jiang, W.~Fan, W.~Li, L.~Wang, Q.~He, P.~Xie, X.~Li, {{DeepFedWT}}: {{A}} federated deep learning framework for fault detection of wind turbines, Measurement 199 (2022) 111529.
\newblock \href {http://dx.doi.org/10.1016/j.measurement.2022.111529} {\path{doi:10.1016/j.measurement.2022.111529}}.

\bibitem{FederatedMulti-ModelTransferLearningBasedFaultDiagnosiswithPeer-to-PeerNetworkforWindTurbineCluster}
W.~Yang, G.~Yu, Federated multi-model transfer learning-based fault diagnosis with peer-to-peer network for wind turbine cluster, Machines 10~(972).
\newblock \href {http://dx.doi.org/10.3390/machines10110972} {\path{doi:10.3390/machines10110972}}.

\bibitem{TowardsFleet-wideSharingofWindTurbineConditionInformationthroughPrivacy-preservingFederatedLearning}
L.~Jenkel, S.~Jonas, A.~Meyer, Privacy-{{Preserving Fleet-Wide Learning}} of {{Wind Turbine Conditions}} with {{Federated Learning}}, Energies 16~(17) (2023) 6377.
\newblock \href {http://dx.doi.org/10.3390/en16176377} {\path{doi:10.3390/en16176377}}.

\bibitem{Communication-EfficientLearningofDeepNetworksfromDecentralizedData}
B.~McMahan, E.~Moore, D.~Ramage, S.~Hampson, B.~A.~y. Arcas, \href{https://proceedings.mlr.press/v54/mcmahan17a.html}{{Communication-Efficient Learning of Deep Networks from Decentralized Data}}, in: A.~Singh, J.~Zhu (Eds.), Proceedings of the 20th International Conference on Artificial Intelligence and Statistics, Vol.~54 of Proceedings of Machine Learning Research, PMLR, {Fort Lauderdale, FL, USA}, 2017, pp. 1273--1282.
\newline\urlprefix\url{https://proceedings.mlr.press/v54/mcmahan17a.html}

\bibitem{hsuMeasuringEffectsNonIdentical2019}
T.-M.~H. Hsu, H.~Qi, M.~Brown, Measuring the {{Effects}} of {{Non-Identical Data Distribution}} for {{Federated Visual Classification}}\href {http://dx.doi.org/10.48550/ARXIV.1909.06335} {\path{doi:10.48550/ARXIV.1909.06335}}.

\bibitem{NonIIDDataSilos}
Q.~Li, Y.~Diao, Q.~Chen, B.~He, Federated {{Learning}} on {{Non-IID Data Silos}}: {{An Experimental Study}}\href {http://dx.doi.org/10.48550/ARXIV.2102.02079} {\path{doi:10.48550/ARXIV.2102.02079}}.

\bibitem{AdaptivePFL}
Y.~Deng, M.~M. Kamani, M.~Mahdavi, Adaptive {{Personalized Federated Learning}}\href {http://dx.doi.org/10.48550/ARXIV.2003.13461} {\path{doi:10.48550/ARXIV.2003.13461}}.

\bibitem{SalvagingLocalAdaptation}
T.~Yu, E.~Bagdasaryan, V.~Shmatikov, Salvaging {{Federated Learning}} by {{Local Adaptation}}\href {http://dx.doi.org/10.48550/ARXIV.2002.04758} {\path{doi:10.48550/ARXIV.2002.04758}}.

\bibitem{HierarchicalClusteringBriggs}
C.~Briggs, Z.~Fan, P.~Andras, Federated learning with hierarchical clustering of local updates to improve training on non-{{IID}} data, in: 2020 {{International Joint Conference}} on {{Neural Networks}} ({{IJCNN}}), {IEEE}, {Glasgow, United Kingdom}, 2020, pp. 1--9.
\newblock \href {http://dx.doi.org/10.1109/IJCNN48605.2020.9207469} {\path{doi:10.1109/IJCNN48605.2020.9207469}}.

\bibitem{Multi-CenterFederatedLearning:ClientsClusteringforBetterPersonalization}
G.~Long, M.~Xie, T.~Shen, T.~Zhou, X.~Wang, J.~Jiang, Multi-center federated learning: {{Clients}} clustering for better personalization, World Wide Web-internet and Web Information Systems 26~(1) (2022) 481--500.
\newblock \href {http://dx.doi.org/10.1007/s11280-022-01046-x} {\path{doi:10.1007/s11280-022-01046-x}}.

\bibitem{Privacy-preservingknowledgesharingforfewshotbuildingenergyprediction:Afederatedlearningapproach}
L.~Tang, H.~Xie, X.~Wang, Z.~Bie, Privacy-preserving knowledge sharing for few-shot building energy prediction: {{A}} federated learning approach, Applied Energy 337 (2023) 120860.
\newblock \href {http://dx.doi.org/10.1016/j.apenergy.2023.120860} {\path{doi:10.1016/j.apenergy.2023.120860}}.

\bibitem{ObjectiveFunctionClustering}
J.~C. Bezdek, Objective function clustering, in: Pattern Recognition with Fuzzy Objective Function Algorithms, {Springer US}, {Boston, MA}, 1981, pp. 43--93.
\newblock \href {http://dx.doi.org/10.1007/978-1-4757-0450-1-3} {\path{doi:10.1007/978-1-4757-0450-1-3}}.

\bibitem{PersonalizationLayers}
M.~G. Arivazhagan, V.~Aggarwal, A.~K. Singh, S.~Choudhary, Federated {{Learning}} with {{Personalization Layers}}\href {http://dx.doi.org/10.48550/ARXIV.1912.00818} {\path{doi:10.48550/ARXIV.1912.00818}}.

\bibitem{PFLHyperNetworks}
A.~Fallah, A.~Mokhtari, A.~Ozdaglar, \href{https://proceedings.neurips.cc/paper_files/paper/2020/file/24389bfe4fe2eba8bf9aa9203a44cdad-Paper.pdf}{Personalized federated learning with theoretical guarantees: A model-agnostic meta-learning approach}, in: H.~Larochelle, M.~Ranzato, R.~Hadsell, M.~Balcan, H.~Lin (Eds.), Advances in Neural Information Processing Systems, Vol.~33, Curran Associates, Inc., Virtual, 2020, pp. 3557--3568.
\newline\urlprefix\url{https://proceedings.neurips.cc/paper_files/paper/2020/file/24389bfe4fe2eba8bf9aa9203a44cdad-Paper.pdf}

\bibitem{FedAvgWithFineTuning}
L.~Collins, H.~Hassani, A.~Mokhtari, S.~Shakkottai, {{FedAvg}} with {{Fine Tuning}}: {{Local Updates Lead}} to {{Representation Learning}}\href {http://dx.doi.org/10.48550/ARXIV.2205.13692} {\path{doi:10.48550/ARXIV.2205.13692}}.

\bibitem{TowardsPFL}
A.~Z. Tan, H.~Yu, L.~Cui, Q.~Yang, Towards {{Personalized Federated Learning}}, IEEE Transactions on Neural Networks and Learning Systems (2022) 1--17\href {http://dx.doi.org/10.1109/TNNLS.2022.3160699} {\path{doi:10.1109/TNNLS.2022.3160699}}.

\bibitem{SurveyPFL}
V.~Kulkarni, M.~Kulkarni, A.~Pant, Survey of {{Personalization Techniques}} for {{Federated Learning}}, in: 2020 {{Fourth World Conference}} on {{Smart Trends}} in {{Systems}}, {{Security}} and {{Sustainability}} ({{WorldS4}}), {IEEE}, {London, United Kingdom}, 2020, pp. 794--797.
\newblock \href {http://dx.doi.org/10.1109/WorldS450073.2020.9210355} {\path{doi:10.1109/WorldS450073.2020.9210355}}.

\bibitem{Serral2011}
E.~Serral, P.~Valderas, V.~Pelechano, Improving the cold-start problem in user task automation by using models at runtime, Information Systems Development\href {http://dx.doi.org/10.1007/978-1-4419-9790-6_54} {\path{doi:10.1007/978-1-4419-9790-6_54}}.

\bibitem{FederatedTransferLearning}
Q.~Yang, Y.~Liu, Y.~Cheng, Y.~Kang, T.~Chen, H.~Yu, Federated {{Transfer Learning}}, {Springer International Publishing}, {Cham}, 2020, pp. 83--93.
\newblock \href {http://dx.doi.org/10.1007/978-3-031-01585-4_6} {\path{doi:10.1007/978-3-031-01585-4_6}}.

\bibitem{penmanshiel}
C.~Plumley, \href{https://doi.org/10.5281/zenodo.5946808}{Penmanshiel wind farm data} (Feb. 2022).
\newblock \href {http://dx.doi.org/10.5281/zenodo.5946808} {\path{doi:10.5281/zenodo.5946808}}.
\newline\urlprefix\url{https://doi.org/10.5281/zenodo.5946808}

\bibitem{kelmarsh}
C.~Plumley, \href{https://doi.org/10.5281/zenodo.5841834}{Kelmarsh wind farm data} (Feb. 2022).
\newblock \href {http://dx.doi.org/10.5281/zenodo.5841834} {\path{doi:10.5281/zenodo.5841834}}.
\newline\urlprefix\url{https://doi.org/10.5281/zenodo.5841834}

\bibitem{edp2016}
\href{https://www.edp.com/en/innovation/open-data/wind-turbine-scada-signals-2016}{Creative commons attribution-sharealike}.
\newline\urlprefix\url{https://www.edp.com/en/innovation/open-data/wind-turbine-scada-signals-2016}

\bibitem{edp2017}
\href{https://www.edp.com/en/innovation/open-data/wind-turbine-scada-signals-2017}{Creative commons attribution-sharealike}.
\newline\urlprefix\url{https://www.edp.com/en/innovation/open-data/wind-turbine-scada-signals-2017}

\bibitem{FederatedOptimizationinHeterogeneousNetworks}
T.~Li, A.~K. Sahu, M.~Zaheer, M.~Sanjabi, A.~Talwalkar, V.~Smith, Federated {{Optimization}} in {{Heterogeneous Networks}}\href {http://dx.doi.org/10.48550/ARXIV.1812.06127} {\path{doi:10.48550/ARXIV.1812.06127}}.

\bibitem{AdaptationStrategiesforAutomatedMachineLearningonEvolvingData}
B.~Celik, J.~Vanschoren, \href{https://api.semanticscholar.org/CorpusID:219573246}{Adaptation strategies for automated machine learning on evolving data}, IEEE Transactions on Pattern Analysis and Machine Intelligence 43 (2020) 3067--3078.
\newline\urlprefix\url{https://api.semanticscholar.org/CorpusID:219573246}

\bibitem{FederatedHyperparameterTuning:ChallengesBaselinesandConnectionstoWeight-Sharing}
M.~Khodak, R.~Tu, T.~Li, L.~Li, M.-F.~F. Balcan, V.~Smith, A.~Talwalkar, Federated hyperparameter tuning: {{Challenges}}, baselines, and connections to weight-sharing, Advances in Neural Information Processing Systems 34 (2021) 19184--19197, \url{https://proceedings.neurips.cc/paper/2021/hash/a0205b87490c847182672e8d371e9948-Abstract.html}.

\bibitem{AComprehensiveSurveyofContinualLearning:TheoryMethodandApplication}
L.~Wang, X.~Zhang, H.~Su, J.~Zhu, A comprehensive survey of continual learning: Theory, method and application, IEEE Transactions on Pattern Analysis \&amp; Machine Intelligence~(01) (5555) 1--20.
\newblock \href {http://dx.doi.org/10.1109/TPAMI.2024.3367329} {\path{doi:10.1109/TPAMI.2024.3367329}}.

\end{thebibliography}

\end{document}